\def\year{2021}\relax
\documentclass[letterpaper]{article} 
\usepackage{aaai21}  
\usepackage{times}  
\usepackage{helvet} 
\usepackage{courier}  
\usepackage[hyphens]{url}  
\usepackage{graphicx} 
\usepackage{natbib}  
\usepackage{caption} 
\usepackage{amsmath}
\usepackage{subcaption}
\usepackage{multirow}
\usepackage{CJKutf8}

\usepackage{color} 

\title{Predicting Geographic Information with Neural Cellular Automata}
\author{
    Mingxiang Chen\textsuperscript{\rm 1},
    Qichang Chen\textsuperscript{\rm 1},
    Lei Gao\textsuperscript{\rm 1},
    Yilin Chen\textsuperscript{\rm 2},
    Zhecheng Wang\textsuperscript{\rm 2}
    \\
}
\affiliations{

    \textsuperscript{\rm 1}Watermirror.ai\\

    KeJiShengTaiYuan 7B, Room 812\\
    Shenzhen, China\\

    \textsuperscript{\rm 2}Stanford University\\
    450 Serra Mall\\
    Stanford, California 94305\\
    
    \textsuperscript{1}{\{chenmingxiang, chenqichang, gaolei, liuzhixin\}@watermirror.ai}, \textsuperscript{2}{\{yilinc2, zhecheng\}@stanford.edu}
}

\begin{document}
\maketitle

\newcommand{\zhecheng}[1]{{\textcolor{black}{#1}}}
\newcommand{\question}[1]{{\textcolor{red}{#1}}}

\begin{abstract}
This paper presents a novel \zhecheng{framework}
using neural cellular automata (NCA) to regenerate and predict geographic information. The model extends the idea of using NCA to generate/regenerate a specific image by training the model with various geographic data, and thus, taking the traffic condition map as an example, the model is able to predict traffic conditions by giving \zhecheng{certain}
induction information. Our research verified the analogy between NCA and gene in biology, while the innovation of the model significantly widens the boundary of possible applications based on NCAs. From our experimental results, the model shows great potentials in its usability and versatility \zhecheng{which are} not available in previous \zhecheng{studies}.
\zhecheng{The code for model implementation is available at https://redacted}.
\end{abstract}

\section{Introduction}

Cellular automata (CA) is a widely used modeling theory. From the perspective of physics, CA refers to a dynamic system defined in a cell space composed of cells with discrete and finite states, which evolved in discrete time dimensions according to certain local rules. Cells are the most basic component of CA which are distributed in discrete Euclidean space positions. Each cell in the lattice grid takes \zhecheng{from a finite set of discrete states},
follows the same local rules of actions, and updates simultaneously according to the rules. Other cells within the local space which may interact with the rules are defined as the ``neighborhood''. While the evolution for each cell only take place based on local information, a large number of cells make the evolution of the entire dynamic system happen through interactions, and hence form a dynamic effect globally. CAs are not determined by strictly defined equations or functions, but are constituted by a series of rules for constructing models. Therefore, CA is a general term for a class of models which is characterized by discrete time, space, and state.

\begin{figure}[b!]
    \centering
    \includegraphics[width=.5\linewidth]{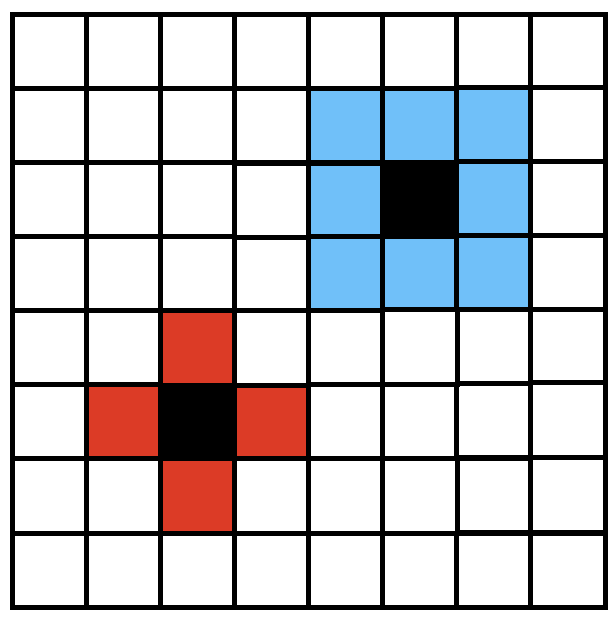}
    \caption{Von Neumann neighborhood (red) and Moore neighborhood (blue).}\label{Fig:intro_neighbour}
\end{figure}

A typical 2-dimensional CA is defined on a finite grid where each square on the grid is a ``cell", which has two possible states ${0,1}$. From another perspective, a cell with a status value of 0 can be considered a dead cell, otherwise it is considered alive. The definition for the ``neighborhood'' of the cell can be varied, for example, \textit{Moore neighborhood} and \textit{von Neumann neighborhood}. A CA starts from a starting pattern called ``configuration" where part of the cells are dead and others are alive, while different configurations and rules may result in various patterns through the simulation. This can be used to simulate real-world scenarios such as pedestrian flow \cite{burstedde2001simulation}\cite{weng2006cellular}, nuclei growth \cite{davies1997growth}, tumour growth \cite{kansal2000simulated}, and many other evacuation process. \zhecheng{It can also be used for generating various visual patterns for aesthetic purposes.}

\begin{figure}[b!]
    \centering
    \includegraphics[width=.95\linewidth]{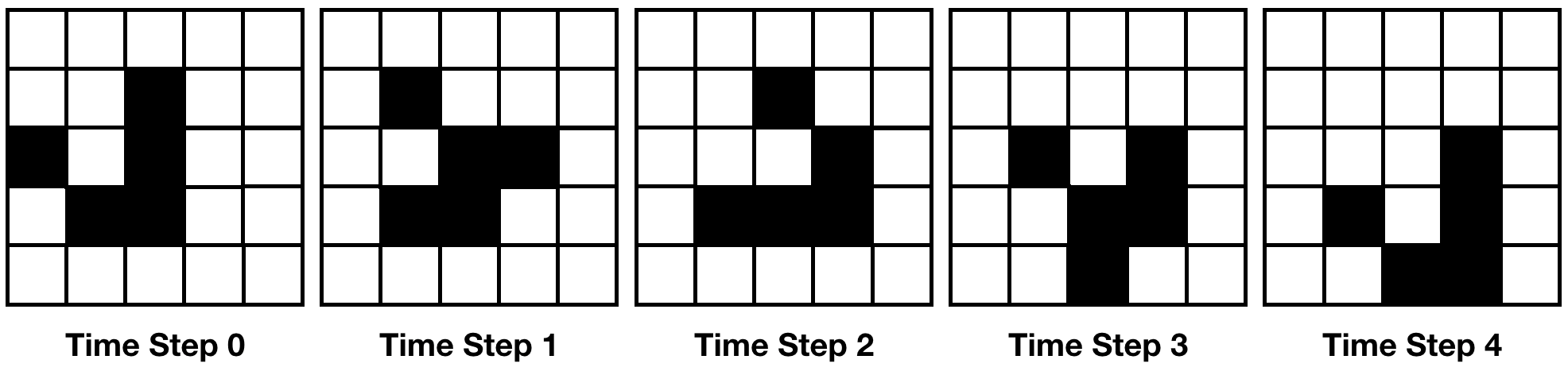}
    \caption{The stages of the glider pattern. Black: alive. White: dead.}\label{Fig:intro_glider}
\end{figure}

\begin{figure}[b!]
    \centering
    \includegraphics[width=.95\linewidth]{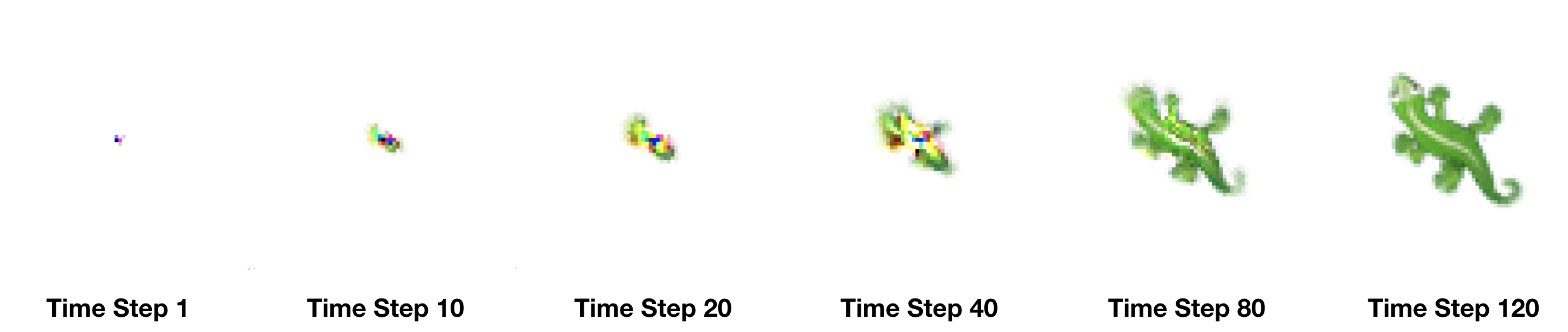}
    \caption{An example of the growing neural cellular automata \cite{mordvintsev2020growing} (our implementation) at different stages.}\label{Fig:NCA_example}
\end{figure}

The earliest research on the combination of neural networks and CA appeared in the late 1980s \cite{chua1988cellular}, and from the beginning of the millennium, many \zhecheng{studies on} 
NCA's applications appeared, such as using NCA for urban planning \cite{li2002neural}\cite{yeh2002urban}\cite{almeida2008using}\cite{qiang2015modeling} and environmental modeling \cite{lauret2016atmospheric}. However, \zhecheng{because of the} 
the constraint of computing power, 
\zhecheng{and the limited capabilities of neural networks at that time,} 
most of these models only use some simple fully connected neural networks. Although they are useful on tasks with certain goals, their versatility and generalization ability still need to be improved. While these shortcomings do not shake their status as the theoretical and experimental cornerstones in this field, the drawbacks do greatly limit the possibility of \zhecheng{converting} 
these studies into \zhecheng{broader} 
applications. With the rapid development of the deep neural network in the past decade, the expression ability of models have also been quickly improved.

In early 2020, the paper \cite{mordvintsev2020growing} from the Google team proposed some new methods related to NCA training. Although the various components of the model are not so special comparing to other deep learning models, after incorporating some biological-related concepts and improving the training method, the final patterns presented by the model \zhecheng{show} 
very amazing effects. A large number of cells together form a stable system, which is very similar to a multi-cellular organism. When a part of the multi-cell system is damaged, using the information of other parts, the damaged area can be quickly restored to the original state. The research team likened this implicit rule learned through deep learning to genes in biology, and the final stable pattern to a multi-cellular organism. They believe that the parameters stored in the NCA model contain hidden information about the growth of the creature's various body parts, while from our perspective, that the obtained pattern can be described as an organ on a creature. Different organs share the same set of genes, but the traits change due to different inductions, which means that rather than growing and repairing one simple emoji, it should be able to learn multiple images so that it can be used to generate new ``organs'' by introducing various inducers just like stem cells. This connection with biology conceptually exhibits Marcavianity \zhecheng{across} 
both time and space. From the perspective of time, the process in which each cell changes according to the same rule to form different patterns can be viewed as a Markov Process, while the parameters controlling each cell's behavior \zhecheng{across space} can be viewed as a Markov random field (MRF). Once we trained the overall network to obtain this MRF, we can generate samples from a local region to restore some kind of broader fields.

\zhecheng{In each step of the growing NCA model as mentioned in the Google's study, each RGBA value in the map are updated with an incremental value $\delta$}.The method of iteratively updating $\delta$ may cause numerical overflow. In this study, we explore the method of iterating a probability distribution to avoid this problem. We also use the above ideas to build a new training method and model \zhecheng{architecture,} 
and take the traffic condition map as an example to show the model’s capability of generating new maps.

This paper is organized as follows. First, we discuss the structure of the model in Section 2. Then the experiments, results, and training details are introduced in Section 3. The conclusion and \zhecheng{potential} 
future works are discussed in the last section.

\section{Model}

\begin{figure*}[!htp]
    \centering
    \includegraphics[width=.8\linewidth]{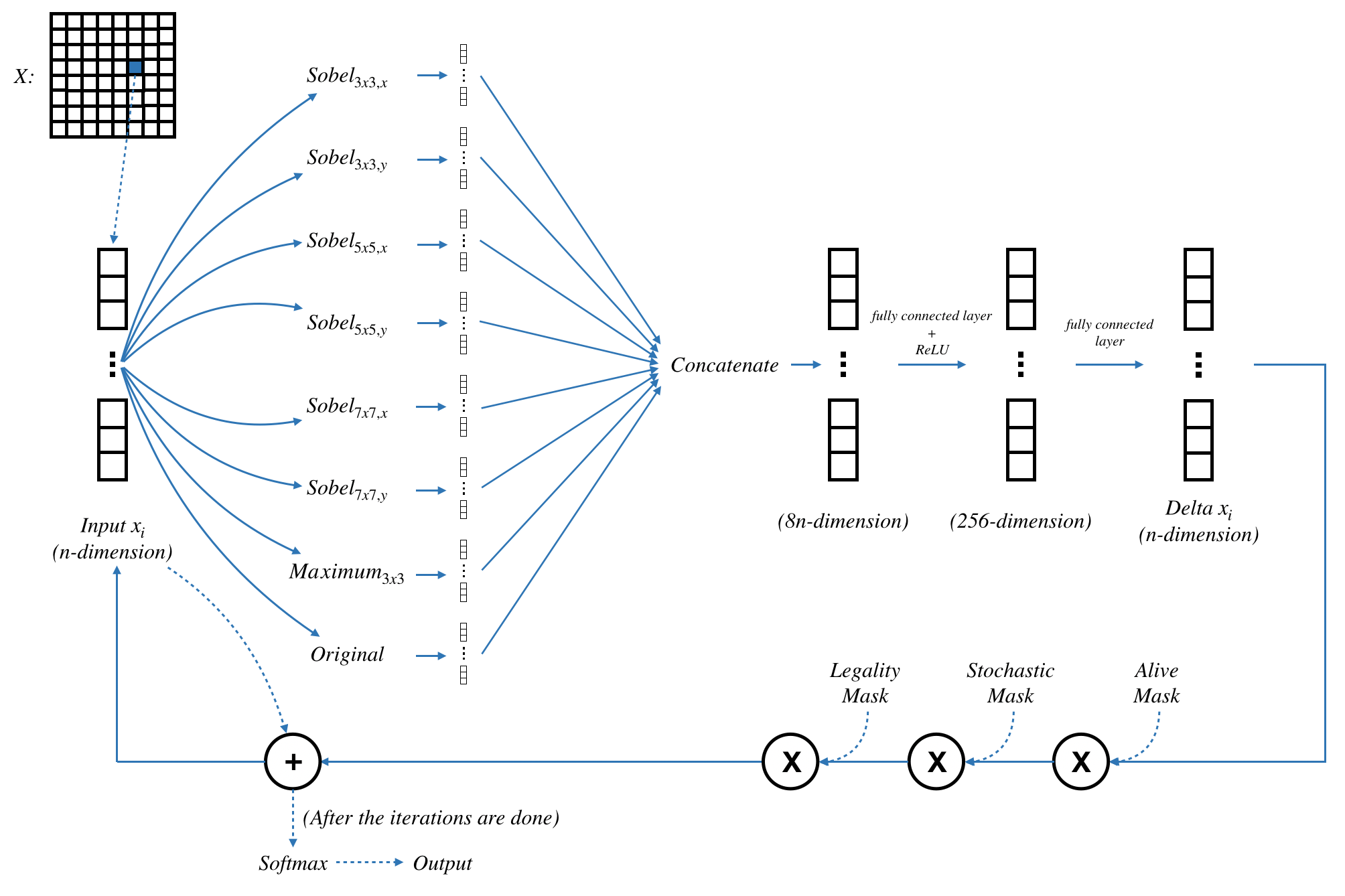}
    \caption{An illustration of the updating process.}\label{Fig:ModelStruc}
\end{figure*}

\subsection{Cell States}

Before introducing the model, let us consider Conway's Game of Life, where each cell has two status: live and dead (or $\{1,0\}$ represented by numbers). The rules controlling this process is rather simple, that any live cell with 2 or 3 alive neighbors survive, and any dead cells with exactly 3 alive neighbors becomes a live cell. Figure 2 is an example pattern named as ``glider'' following this rule. While we would like to enlarge the information that a single cell could carry, for example, increase the number of status, so that a more complex pattern can be represented \zhecheng{across} 
the cellular space. Following the paradigm given in the previous research \cite{mordvintsev2020growing}, the status for each cell could be represented by \zhecheng{a vector instead of 0 or 1}, 
where the values \zhecheng{in the vector} can be used to describe the ratio of different elements, such as RGBA values, or other hidden information.

In this paper, we define the input tensor, which has a shape of (height, width, $n$), as a ``map'', and every single vector of length $n$ \zhecheng{at each location} on this map as a ``cell". Rather than colors, the outputs of the model is a set of probabilities, so that the first $k$ dimensions in a cell represent the logits, where the probabilities will then be calculated by the $Softmax$ function. The $(k+1)^{th}$ dimension represents the ``aliveness''. \zhecheng{I.e. the cell is alive if the probability is above 0.1 and dead otherwise.} 
The rest \zhecheng{of the} dimensions contain the hidden values which quicken the training process. As stated by other researchers, these values can be interpreted as ``concentrations of some chemicals, electric potentials, or some other signaling mechanism that are used by cells to orchestrate the growth" \cite{mordvintsev2020growing}. The starting states, or the ``configuration" of our NCA is simply zero out all cell states, except a cell at a random location, where that value of both its $(k+1)$th dimension \zhecheng{(``aliveness'')} and hidden values are set to 1.

\subsection{Update Rules}

In each step, the model takes the values of the entire map and outputs $\delta_{update}$ at time step $t$, where the values for each cell is updated by:

\begin{equation}
    x_{i,t} = x_{i,t}+\beta*\delta_{update,i,t}\\
\end{equation}

In order to obtain information at different scales, each cell applies sobel filters to its surrounding 3 by 3, 5 by 5, and 7 by 7 squares. Besides, the maximum value within the surrounding 3 by 3 area is also calculated. The above information is then concatenated with the vector representing the cell itself, and then through two fully connected layers, the $\delta_{update}$ value of each cell is obtained. However, it is worth noting that although each cell can calculate a $\delta_{update}$ value, not all cells will be updated. In this model, three types of masks are used: alive mask, stochastic mask, and legality mask.
Firstly, a cell can be updated only if there is at least one living cell in its surrounding area, while the size of the ``surrounding area'' may be varied for different tasks \zhecheng{which is imposed by the alive mask}. Secondly, the stochastic mask gives randomness to updating values \zhecheng{in order to} improve the robustness of the model. Thirdly, the legality mask indicates which part of the map is valid for a given task. For instance, the legality mask may be the available land areas on the given map. The predefined probability of the stochastic mask for all the experiments in this paper is 0.5.

The output provided by the model after $T$ iterations are the logits (first k dimensions), aliveness (the $(k+1)$th dimension) and the hidden values. The classification results are then extracted by taking a $Softmax$ function on the logits if the cell is alive.

\subsection{Pre-Explored Areas}

Common prediction tasks usually consist of input data, models, predicted values, and the ground truth values. However, as mentioned earlier, our model aims to mimic the process of stem cell differentiation. \zhecheng{Given} 
a certain induction, even if carrying the same set of genes, the stem cells 
\zhecheng{can be} 
induced to differentiate into organs with various functions. In this research, the part where the induction is carried out is called ``pre-explored areas'', which means that part of the map is already known at the initial stage. When the ground truth probabilities and \zhecheng{$\alpha$} 
values for some of the cells are known, they would act like inducing for cell differentiation in biology. However, since the probability for each class is calculated from the $Softmax$ function, it is impossible to reversely derive the logits. Inspired by the constant stimulation \zhecheng{which} can change some biological characteristics of animals and plants, such as the phototaxis of plants, and the treatment to correct teeth arrangement, we apply an additional gradient descent to each cell within the pre-explored area at every step that

\begin{equation}
    x_{i,t} = x_{i,t}+\beta*\delta_{update,i,t}-C*\delta_{pre,t}\\
\end{equation}

\begin{align}
    \delta_{pre,t} & = \frac{\partial KL(p,h)}{\partial x_{i,t}} \\
                   & = \frac{\partial p_i\log \frac{p_i}{h_{i,t}}}{\partial x_{i,t}} \\
                   & = -p_i(1-h_{i,t})
\label{eq:3}
\end{align}
where $x_{i,t}$, $h_{i,t}$, and $p_i$ are the logit, predicted probability, and ground truth probability for the $i^{th}$ cell at the $t^{th}$ time step, respectively, and $C$ is the hyper-parameter representing the concentration of induction. The cell will quickly \zhecheng{converge to} 
a probability very similar to the ground truth values after a few iterations.

\subsection{Objectives}

We train the model so that after T iterations, the \zhecheng{state} 
generated by sampling the probability distribution output by the model is as close as possible to the ground truth values. Therefore, we use the KL divergence to represent the difference between the predicted probability distribution and the actual, and the square loss to calculate the error of cell viability:

\begin{align}
    loss & = \sum_{i} loss_{i,KL} + loss_{i,viability} \\
         & = \sum_{i} KL(h_{i,T}, p_i) * alive_{i} + (\alpha_{i,T}-alive_{i})^2
    \label{eq:2}
\end{align}
where $a_{i,T}$ is the alpha value (the $(k+1)^{th}$ element of $x_{i,T}$), and $alive_i$ represents the ground truth value of whether the cell should be alive or not. The hidden values will not be used to calculate the loss.

\section{Applications}

To verify the performance of the model we designed under some image prediction tasks, we designed two sets of experimental schemes based on traffic maps. These traffic condition maps are acquired using the web service API from amap.com\footnote{https://lbs.amap.com/api/webservice/summary/}. We collected 1126 traffic maps of ten locations from Shenzhen, a city located at southern China, at different time points from 6 AM to 11 PM. The traffic conditions at midnight and early morning are very smooth, so in order to avoid the training data being too similar, these data are not collected. The traffic maps are square areas with a side length of 1.6 km. Each map has a \zhecheng{size} 
of 80 by 80, so that each pixel represents a square with \zhecheng{a side length} 
of 20 meters. When collecting the data, the web API provides a total of five traffic conditions: unobstructed (green), slightly congested (yellow), moderately congested (orange), severely congested (red), and extremely congested (purple). \zhecheng{Since} 
the last two categories, especially the extremely congested condition, is too sparse, the two channels are merged into one represented by the red color in our experiments.

\begin{CJK*}{UTF8}{gbsn}

\begin{table}[!t]
\centering
\begin{tabular}{|c|c|}
\hline
    \begin{tabular}{c}
        Name (Number of\\
        Maps)
    \end{tabular} &
    \begin{tabular}{c}
        Location
    \end{tabular}\\
    \hline
    
    \begin{tabular}{c}
        深圳湾科技生态园\\
        KeJiShengTaiYuan (270)
    \end{tabular} &
    \begin{tabular}{c}
        113.9527, 22.5304
    \end{tabular} \\
    \hline
    
    \begin{tabular}{c}
        来福士广场\\
        LaiFuShi Square (92)
    \end{tabular} &
    \begin{tabular}{c}
        113.9258, 22.5120
    \end{tabular} \\
    \hline
    
    \begin{tabular}{c}
        福田区市民中心\\
        Citizen Center (94)
    \end{tabular} &
    \begin{tabular}{c}
        114.0596, 22.5437
    \end{tabular} \\
    \hline
    
    \begin{tabular}{c}
        宝安区海雅缤纷城\\
        HaiYaBinFenCheng (108)
    \end{tabular} &
    \begin{tabular}{c}
        113.9061, 22.5597
    \end{tabular} \\
    \hline
    
    \begin{tabular}{c}
        深圳科学馆\\
        ShenZhen Science \\
        Museum (96)
    \end{tabular} &
    \begin{tabular}{c}
        114.0961, 22.5424
    \end{tabular} \\
    \hline
    
    \begin{tabular}{c}
        红山地铁站 \\
        HongShan\\
        Subway Station (96)
    \end{tabular} &
    \begin{tabular}{c}
        114.0226, 22.6522
    \end{tabular} \\
    \hline
    
    \begin{tabular}{c}
        光明行政服务大厅\\
        GuangMing Admini- \\
        strative Service Hall (109)
    \end{tabular} &
    \begin{tabular}{c}
        114.0424, 22.6692
    \end{tabular} \\
    \hline
    
    \begin{tabular}{c}
        海上世界\\
        Sea World (76)
    \end{tabular} &
    \begin{tabular}{c}
        113.9151, 22.4851
    \end{tabular} \\
    \hline
    
    \begin{tabular}{c}
        华侨城小学\\
        HuaQiaoCheng \\
        Elementary \\
        School (103)
    \end{tabular} &
    \begin{tabular}{c}
        113.9868, 22.5366
    \end{tabular} \\
    \hline
    
    \begin{tabular}{c}
        车公庙地铁站\\
        CheGongMiao \\
        Subway Station (82)
    \end{tabular} &
    \begin{tabular}{c}
        114.0255, 22.5348
    \end{tabular} \\
    \hline
\end{tabular}

\caption{The locations (represented by (longitude, latitude)) of the traffic condition maps}\label{table: traffic map positions}
\end{table}

\end{CJK*}

In the first task, only the traffic maps of the first location (KeJiShengTaiYuan) are used for training and testing (270 maps in total, 245 for training and 25 for testing). The goal of this task is to explore the model's ability to predict traffic conditions in the same area at different times.

As for the second task, the traffic conditions at all locations are used for training and testing. Among them, each location provides 64 randomly selected maps and forms the training set, while the remaining maps become the test set. In this task, the goal is to use data from different locations to train a universal NCA of traffic prediction. For the training set and the test set in both tasks, the traffic conditions in the pre-explored area, which is a random circular area whose diameter is half the side length, is known in advance.

\section{Experiments and Results}

\begin{figure}[!t]
    \centering
    \includegraphics{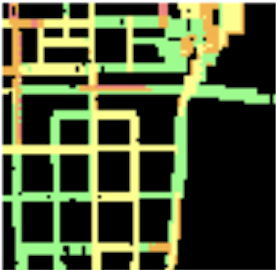}
    \caption{An example of traffic condition map in Shenzhen KeJiShengTaiYuan.}\label{Fig:consecutive}
\end{figure}


In the second section of this paper, we talked about the model's \zhecheng{architecture,}
where the cells are firstly parsed by filters implemented by the depth-wise kernels. As indicated in the previous research \cite{mordvintsev2020growing}, motivated by the chemical gradient signals in real-life creatures, the Sobel filter shows its good performance in building rules for NCAs. However, unlike images, geographic information maps, often contain a large number of consecutive areas with similar information. As shown in figure \ref{Fig:consecutive}, areas often appear as large patches of the same color on maps, which makes it difficult for the model to reason about the boundaries. To solve this problem, we extend the idea of using Sobel filters by adding more filters of various types and sizes into the neural network. Moreover, in addition to logits and the alpha value, each cell also carries a certain amount of hidden information to simulate chemical or electrochemical stimulation in an organism, which is often difficult to observe directly, but is critical to the operation of biological machines. From another point of view, vectors in high-dimensional space can carry more complex signals, making the interaction between cells more efficient. It is worth noting that the additional gradient descent step for the cells within the pre-explored area is only applied to the first $k$ dimensions, and the aliveness and the hidden values would not be manipulated.

\begin{figure}[!b]
    \centering
    \begin{subfigure}[b]{0.99\linewidth}
        \includegraphics[width=.99\linewidth]{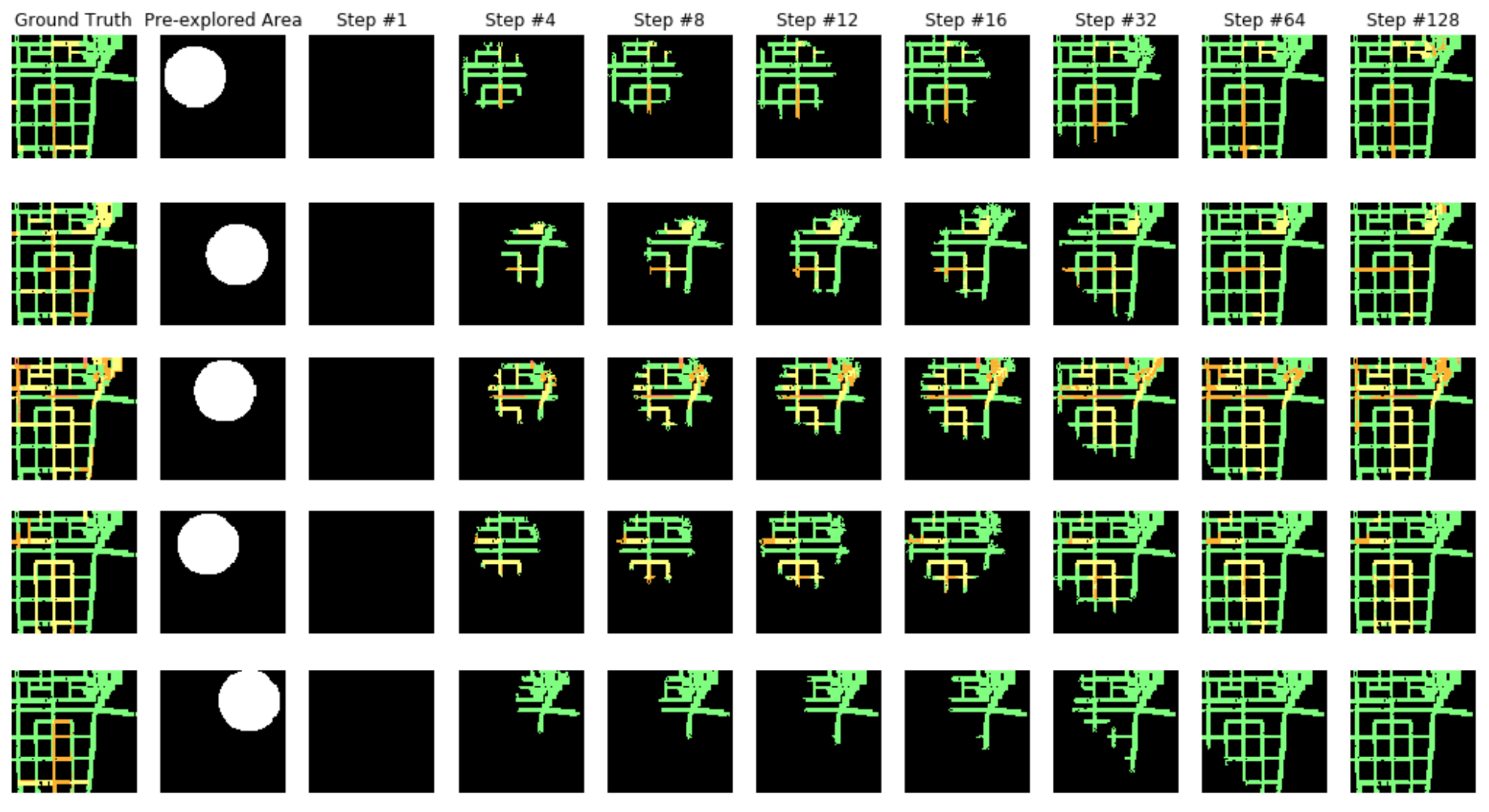}
        \caption{Growing}
        \label{fig:TrafficResult2_0} 
    \end{subfigure}
    \begin{subfigure}[b]{0.99\linewidth}
        \includegraphics[width=.99\linewidth]{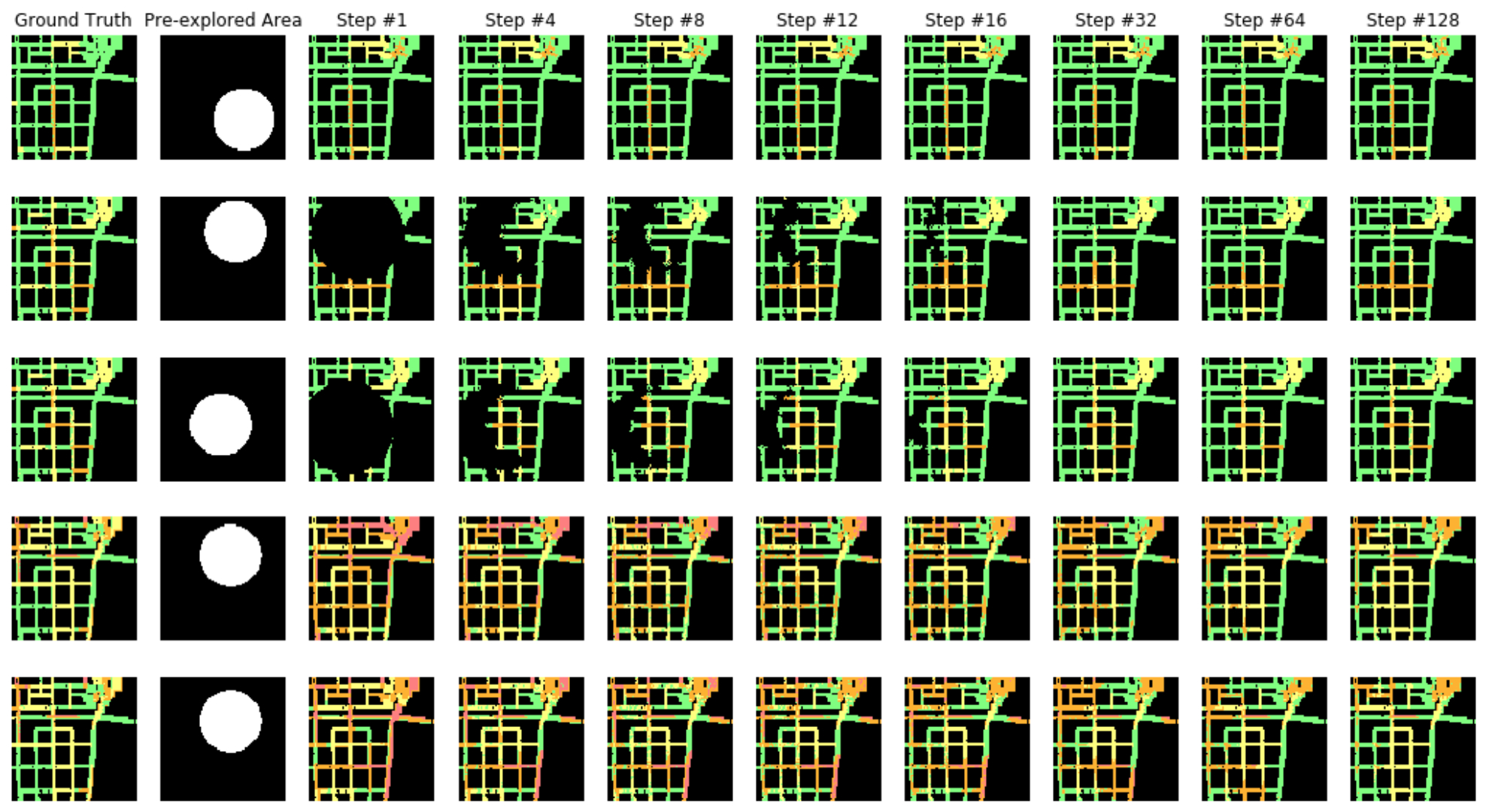}
        \caption{Persisting, regeneration, and transformation}
        \label{fig:TrafficResult2_1} 
    \end{subfigure}
    \caption{An illustration of traffic prediction at the same location of different time. The figure on the top shows how the model "grow" a traffic map from a pre-explored area. The figure on the bottom shows how the model "persists" (row 1), "regenerates" (row 2 \& 3), and "transform" (row 4 \& 5) a traffic map to the target if part of it is pre-explored. In these diagrams, the first column from the left is the ground truth traffic map, and the second column shows the pre-explored area.}\label{Fig:TrafficResult}
\end{figure}

\begin{figure}[!b]
    \centering
    \includegraphics[width=.99\linewidth]{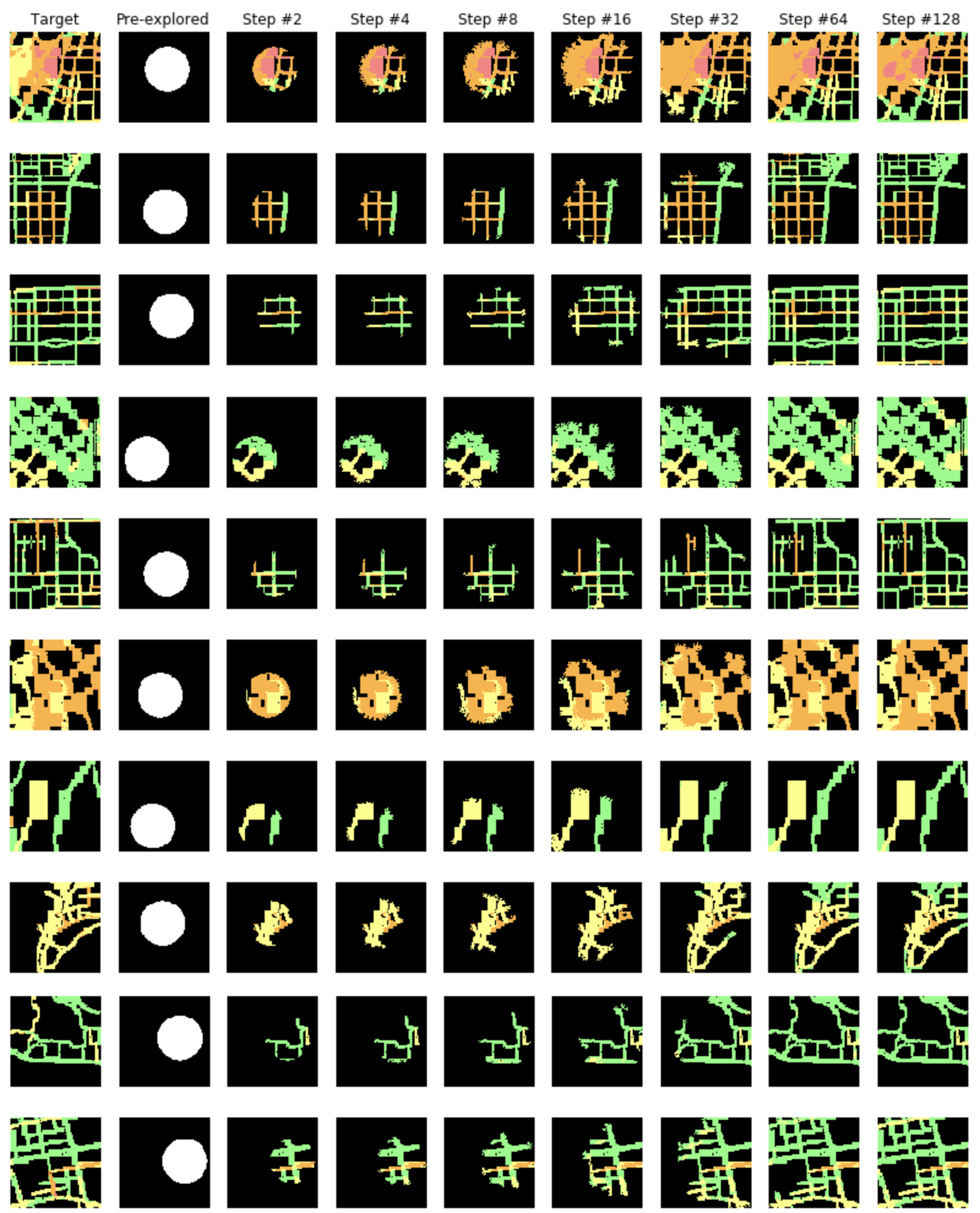}
    \caption{An illustration of traffic prediction of multiple locations at different time. The first column from the left shows the ground truth traffic map, and the second column shows the pre-explored area. The rest illustrates how the CA iterates at different time steps. The result demonstrates that the model can be generalized to different maps, and  predict the traffic by only giving a part of the traffic condition in the certain area.}\label{Fig:MultiTrafficResult}
\end{figure}

Considering the balance between the computation speed and accuracy, each cell is composed of a 16-dimensional vector, where the first $k=4$ dimensions are logits of the four traffic situations, the fifth dimension represents the aliveness, and the rest of the dimensions are hidden values. Based on the above, the input tensors have a shape of (80, 80, 16). 

\begin{table}[!b]
    \centering
    \begin{tabular}{|c|cc|c|}
    \hline
        & \multicolumn{2}{c}{Accuracy} & \\
        Task & Train & Test & Time (s)\\
        \hline
        Mono-location Traffic & 79.08\% & 81.26\% & \multirow{2}{*}{0.0412} \\
        Multi-location Traffic & 74.15\% & 76.31\% & \\
    \hline
    \end{tabular}
    \caption{Performance 
    on the training set and the test set. Both tasks share the same model, that only one average inference time (per instance running 128 steps) is given.}
    \label{tab:performance}
\end{table}

For the first task, the result on the test set can be seen from figure \ref{Fig:TrafficResult}, where we set four different tasks \zhecheng{when}  
training: 1. growing a map from pre-explored areas; 2. persisting the prediction along with iterations; 3. regenerate the traffic condition if part of it is missing; 4. transform from an incorrect map to a correct one. The result demonstrates the hypothesis that the ``gene" can be generalized such that the output can not only be interpreted as a ``creature", but also an ``organ", and use the ``gene" together with some ``induction" (pre-explored area) to guide the growth of cells and tissues. For the second task, the model also shows a good performance on predicting the traffic conditions as illustrated in figure \ref{Fig:MultiTrafficResult}, which further shows the usability and versatility of the model.

The first task in this section is trained for 50,000 epochs using Adam optimizer which costs less than 9.5 hours on a TITAN RTX GPU, and the second model is trained for 100,000 epochs costing 18 hours using the same hardware. The number of iterating steps $T$ in both tasks is 128. During both training and testing, the values in a cell would be considered as ``updatable" only if at least one of its adjacent cells (the surrounding 3 by 3 area) is alive. The performance 
for the above two examples can be found in table \ref{tab:performance}. The accuracy in the table is defined as

\begin{equation}
    accuracy=\frac{|T \cap P|}{|P|}\\
\end{equation}
where P and T are sets of predicted labels and true labels, respectively. The results are averaged among all training or testing instances, where for each instance, we tested 10 times with different random located pre-explored areas. Since the test set for each location in the multi-location traffic prediction task is \zhecheng{imbalanced,} 
the average accuracy for each location will be calculated separately, and then the overall average accuracy will be calculated based on these results. It shows that our algorithm achieves a high performance both on the training set and the test set, while the inference time is also \zhecheng{short.} 
Nevertheless, there are still some mislabeled pixels. For instance, the last row of figure \ref{fig:TrafficResult2_0} indicates the prediction may go wrong if improper pre-explored area is given. Since all the pixels pre-explored in that map are green, the model can hardly output the correct traffic map.

\section{Conclusion}

The purpose of this study is to explore a way to combine CA and neural networks, and hence modeling the geographic information. Compared with previous studies, the outcome has a stronger generalization ability which means the model can be adapted to multiple geographic maps rather than one image only, and furthermore, predicting and restoring geographic data. Based on the results of this study, the combination of neural network and CA may be a promising idea and may create its value in more aspects that are closely related to human activities.

\bibliography{main}

\end{document}